\documentclass[11pt]{article}
\usepackage[utf8]{inputenc}
\usepackage{graphicx}
\usepackage{amsmath,amssymb}
\usepackage{geometry}
\usepackage{natbib}
\usepackage{hyperref}
\usepackage{array}
\title{DYNARTmo: A Dynamic Articulatory Model for Visualization of Speech Movement Patterns}
\author{Bernd J. Kröger\\
\small Medical School, RWTH Aachen University, Aachen, Germany\\
\small Kröger Lab, Belgium, \url{www.speechtrainer.eu}}
\date{}

\begin{document}

\maketitle

\begin{abstract}
We present DYNARTmo, a dynamic articulatory model designed to visualize speech articulation 
processes in a two-dimensional midsagittal plane. The model builds upon the UK-DYNAMO 
framework and integrates principles of articulatory underspecification, segmental and gestural 
control, and coarticulation. DYNARTmo simulates six key articulators based on ten continuous and 
six discrete control parameters, allowing for the generation of both vocalic and consonantal 
articulatory configurations. The current implementation is embedded in a web-based application 
(SpeechArticulationTrainer) that includes sagittal, glottal, and palatal views, making it suitable for 
use in phonetics education and speech therapy. While this paper focuses on the static modeling 
aspects, future work will address dynamic movement generation and integration with articulatory-acoustic modules.
\end{abstract}

\section*{1. Introduction}
Articulatory models describe the stage of speech where neuromuscular activation patterns lead to 
movement patterns of speech articulators (i.e., lips, tongue, velum, lower jaw) with the goal of 
generating audible speech signals as part of speech communication processes. The motivation for 
developing articulatory models is manifold. (i) Models unfold the process of speech articulation 
and acoustic signal generation and thus help to understand the basic mechanisms of articulation 
and coarticulation. (ii) If coupled with an articulatory-acoustic module, an articulatory model 
provides insight into the mechanisms of acoustic signal generation at the vocal folds and within 
the vocal tract. (iii) Articulatory models may serve in the future as high-quality speech synthesizers, 
capable of generating different voices (i.e., different speakers), different emotional connotations, 
and more—beyond merely transmitting a spoken message. However, corpus-based synthesis systems 
currently produce the most natural speech quality, and many research questions remain to be addressed 
before high-quality articulatory synthesizers can be realized \citep{Campbell2005, Kroger2022}.

At present, articulatory models are mainly used as research tools to study articulation and 
coarticulation or, when combined with articulatory-acoustic modules, to investigate aspects of 
acoustic signal generation, including sublaryngeal, laryngeal, and supralaryngeal 
aerodynamics \citep{Birkholz2013, Kroger2022, Fan2024}. Other models are used in speech therapy 
to visually present articulator positions (\citep{Kroger2005}) 
or articulatory contact information (e.g., tongue–palate or labial contact), 
offering somatosensory feedback to help improve our imagination of articulation. 
Examples include electropalatography (EPG), which provides real‑time visualization 
of tongue–palate contact during speech (e.g. \citep{Hardcastle1991, Nordberg2011}), 
and ultrasound visual biofeedback (U‑VBF), displaying dynamic tongue surface movements in 
therapy for speech sound disorders (\citep{Sugden2019, Preston2017}). Moreover, 
articulatory models 
can be used to simulate speech learning or relearning processes in therapeutic 
contexts, or as part of second and third language learning. In that case, articulatory models 
need to be integrated into large-scale neurobiologically motivated neural models of language 
processing \citep{Kroger2023}.

Many aspects of speech articulation can be represented in the midsagittal plane. Therefore, early 
articulatory models were typically two-dimensional \citep{Henke1966, Mermelstein1973, Coker1976, 
Maeda1979, Heike1979}. These models primarily aimed at generating high-quality synthetic speech. 
However, by the end of the 20th century, corpus-based synthesizers such as the Klatt synthesizer 
\citep{Klatt1980} and newer systems like corpus-bases speech synthesizers \citep{Campbell2005} 
had outperformed articulatory models in speech quality. Consequently, articulatory models---including 
articulatory-acoustic variants---became more focused on their use as research tools, particularly 
in studying speech acquisition, articulation, and coarticulation across languages and speaker groups. 
As a result, increasingly detailed models emerged that incorporate 3D spatial information, 
neuromuscular control schemes, and biomechanical modeling of muscles and articulatory tissues 
\citep{Kroger2022}.

The model described in this paper, DYNARTmo (DYNamic ARTiculatory model), is a further development 
of the UK-DYNAMO (Universität Köln-DYNamic Articulatory MOdel) originally introduced by Heike 
\citep{Heike1979}. The UK-DYNAMO was inspired by Henke's approach \citep{Henke1966} and later used in a 
computer-based tool for speech training \citep{Heike1989}. Like many models of its time, UK-DYNAMO 
was a 2D system based on midsagittal contours of articulator surfaces, specifically German vowel and 
consonant contours from Wängler \citep{Waengler1958}. It employed temporal interpolation and a 
basic form of coarticulation as suggested by Henke.

DYNARTmo was initially introduced by Kröger \citep{Kroger1992}, who incorporated the concept of 
articulatory underspecification to derive minimal rules for articulation and coarticulation. 
This approach was elaborated through two control concepts: a gestural control scheme \citep{Kroger1993, 
KrogerEtAl1995} and a segmental approach \citep{Kroger2001}. Starting in the 2000s, DYNARTmo was 
updated with geometries derived from MRI data \citep{KrogerEtAl2000, KrogerEtAl2004}, offering a more 
realistic articulatory basis. Later, it was embedded into a large-scale neurobiological speech 
production model \citep{KrogerEtAl2014}, incorporating concepts such as primary and secondary 
articulators, muscle bundles, and neuromuscular activation patterns \citep{KrogerBekolay2022}.

This paper presents the current implementation of DYNARTmo as part of the SpeechArticulationTrainer 
web application, available at \url{https://speechtrainer.eu/projects.htm}. In addition to the 
midsagittal view, the app now includes glottal (superior) and palatal (inferior) views. Together, 
these three perspectives offer an integrated spatial understanding of articulatory processes. 
While the focus of this paper is on the static aspects of the model, future work will address 
the generation of articulatory movement patterns.

\section*{2. DYNARTmo: Static Aspects}

\begin{figure}[ht]
    \centering
    \includegraphics[width=\textwidth]{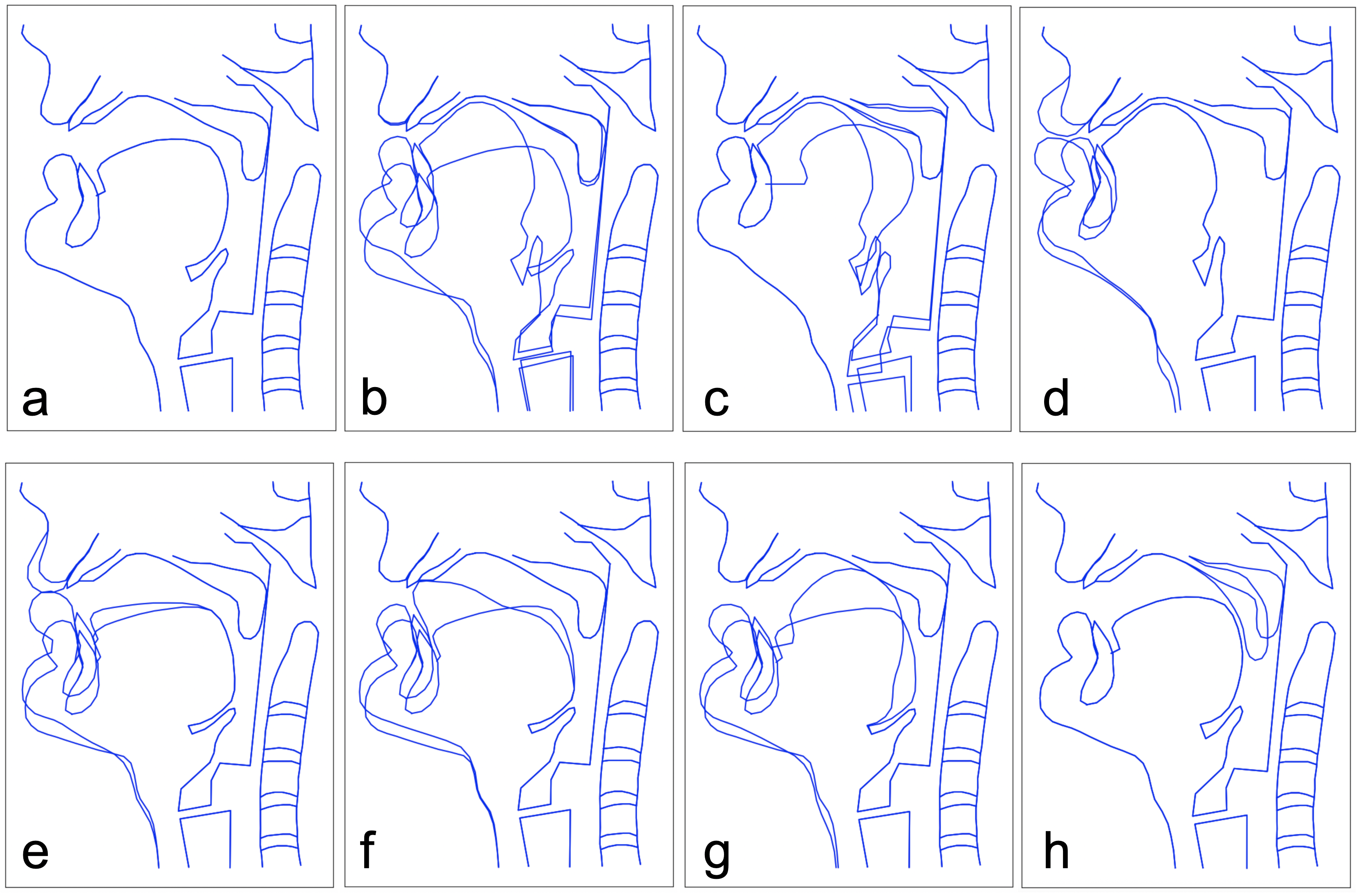}
    \caption{Midsagittal views generated by DYNARTmo. (a) shows the neutral configuration of all 
    articulator contours for a central vocalic state. (b--h) display paired overlays of articulator 
    contours, illustrating edge positions (displacements) for individual control parameters: 
    (b) vocalic height (high--low) for a front vowel position; (c) vocalic front--back for a high 
    vowel position; (d) lip rounding (spread--rounded) for a high front vowel; (e) labial aperture 
    from vocalic to full closure (for a low vowel); (f) tongue tip height from vocalic to 
    closure (for a low vowel); (g) tongue dorsum height from vocalic to closure (for a low vowel); 
    (h) velum height (velopharyngeal aperture: closed--open) for the neutral vocalic position.}
    \label{fig:static_contours}
\end{figure}

DYNARTmo simulates six model articulators---lips, tongue tip, tongue dorsum, lower jaw, velum, and 
glottis---which are controlled by a set of ten continuously scaled control parameters 
(see Table~\ref{tab:control_parameters} and Figure~\ref{fig:static_contours}). These control 
parameters govern the displacement of each articulator from its neutral or rest position. 
Functionally, the parameters are grouped into three vocalic parameters, three consonantal parameters, 
one parameter for velum control, and three parameters for phonatory control. It is important to note 
at this point that the lower jaw is not directly controlled by any of the parameters introduced here 
(see \textit{Discussion}).

The vocalic control parameters specify tongue height (high–low), tongue position along the front–back 
axis, and the degree of lip rounding (spread–rounded). The consonantal parameters define the degree 
of constriction generated by the lips, tongue tip, or tongue dorsum. These values are relative to 
the current vocalic articulator position and describe, for example, the elevation of the tongue tip 
or dorsum toward closure, or respectively the reduction of labial aperture from an open vocalic state 
to full closure.

In addition, further control parameters are defined for the velopharyngeal and laryngeal--sublaryngeal 
systems (Table~\ref{tab:control_parameters}). Velum height controls the opening of the velopharyngeal 
port. Glottal aperture sets the position of the arytenoid cartilages, thereby determining the resting 
distance of the vocal folds for breathing, voiceless sounds, or phonation. Vocal fold tension controls 
the pitch (i.e., the fundamental frequency) of phonation. Lung pressure governs the production of 
airflow required for phonation and determines the loudness (intensity) of the resulting speech signal.

\begin{table}[ht]
\centering
\caption{Continuously scaled control parameters used in DYNARTmo, including affected articulators 
and their functional roles. Vocalic parameters define global vocal tract shaping; consonantal parameters 
govern localized constrictions. Additional parameters control velopharyngeal aperture, 
glottal configuration, and subglottal pressure. (Speech sounds are transcribed using 
SAMPA; see \citep{Wells1997})}
\label{tab:control_parameters}
\renewcommand{\arraystretch}{1.3}
\begin{tabular}{ll>{\raggedright\arraybackslash}p{4.2cm} >{\raggedright\arraybackslash}p{6.0cm}}
\hline
\textbf{Type} & \textbf{Name} & \textbf{Affected Articulators} & \textbf{Function} \\
\hline
voc   & vocalic high--low          & tongue body, lower jaw           & vertical shaping of the vocal tract (vowel height) \\
voc   & vocalic front--back        & tongue body, lower jaw           & horizontal shaping of the vocal tract (vowel position) \\
voc   & lips spread--round         & lips                             & lip rounding, relevant for vowels and some consonants (e.g., /S/) \\
cons  & labial aperture            & lips, lower jaw                  & constriction for labial consonants (e.g., /b/, /p/, /m/) \\
cons  & tongue tip height          & tongue tip, tongue body, lower jaw & constriction in the apical region (e.g., /d/, /t/, /n/) \\
cons  & tongue dorsum height       & tongue dorsum, tongue body, lower jaw & constriction in the dorsal region (e.g., /g/, /k/, /N/) \\
vel   & velum height               & velum                            & degree of velopharyngeal opening (nasality control) \\
glott & glottal aperture           & arytenoid cartilages (larynx)    & voicing control: open for voiceless, closed for phonation \\
glott & vocal fold tension         & vocal folds (larynx)             & pitch (fundamental frequency) control via tension modulation \\
lung  & lung pressure              & sublaryngeal system              & controls airflow intensity and loudness of the speech signal \\
\hline
\end{tabular}
\end{table}

In addition to the ten continuously scaled control parameters (see Table~\ref{tab:control_parameters}), 
the model includes six discrete parameters that define the place and manner of consonantal constrictions 
(see Table~\ref{tab:discrete_parameters}). The value range of the continuous parameters varies 
depending on their functional category: for vocalic parameters, values range from $-1$ to $1$, 
where $0$ indicates the neutral articulator position. The same $-1$ to $1$ scale is used for 
velopharyngeal and glottal aperture, with $1$ indicating full opening, $0$ indicating closure, 
and $-1$ representing tight closure. For parameters governing consonantal constrictions (i.e., 
labial, apical, and dorsal), as well as for vocal fold tension and lung pressure, the scale ranges 
from $0$ to $1$. A value of $0$ indicates no active constriction or minimum effort 
(i.e., the articulator remains in its current vocalic position), while a value of $1$ represents 
full articulatory closure or maximum effort.

The discrete control parameters (Table~\ref{tab:discrete_parameters}) define both the \emph{place} 
and \emph{manner} of consonantal articulation. The manner of articulation distinguishes between: 
(i) full closure (as in plosives and nasals), (ii) fricative closure, in which the airstream is 
focused through a midsagittal groove (or, in labiodental fricatives, between the lower teeth and 
upper lip), and (iii) lateral closure, which involves strong central raising and simultaneous 
lateral lowering of the tongue surface, as in laterals. These manners correspond to distinct 
articulatory configurations or shapes. 

The place of articulation is categorized according to anatomical regions: \emph{bilabial} 
or \emph{labiodental} for the lips; \emph{dental}, \emph{alveolar}, or \emph{post-alveolar} for 
the tongue tip; and \emph{palatal} or \emph{velar} for the tongue dorsum. It should be noted that 
although tongue tip and tongue dorsum positions are represented by continuously scaled parameters 
in the current model implementation, they are typically assigned discrete values to reflect 
linguistically relevant articulatory targets during consonant production.

\begin{table}[ht]
\centering
\caption{Discrete control parameters in DYNARTmo, defining place and manner of consonantal constriction. Each articulator is associated with specific labels for place and manner of articulation. Manner defines the constriction shape; place defines the anatomical target region.}
\label{tab:discrete_parameters}
\begin{tabular}{lll}
\hline
\textbf{Articulator} & \textbf{Parameter Type} & \textbf{Possible Labels} \\
\hline
lips         & place  & bilabial, labiodental \\
lips         & manner & full (vocalic), near (central groove or lower teeth row) \\
tongue tip   & place  & dental, alveolar, post-alveolar \\
tongue tip   & manner & full (vocalic), near (central groove), lateral \\
tongue dorsum & place & palatal, velar \\
tongue dorsum & manner & full (vocalic), near (central groove) \\
\hline
\end{tabular}
\end{table}

The set of articulatory control parameters does not directly determine the positions of the model 
articulators. Rather, these parameters serve a functional role: they specify consonantal articulation 
in terms of the degree and location of vocal tract constrictions, and vocalic articulation in terms 
of canonical phonetic-linguistic categories such as high--low, front--back, and spread--rounded 
(cf.\citep{IPA1999}: International Phonetic Association).

The actual specification of consonantal articulator positions is carried out in a subsequent computational step, 
based on the values of all control parameters at a given time point. This step involves computing 
the vertical and horizontal positions of the tongue tip, tongue dorsum, and lips, as well as the 
vertical positions of the velum, lower jaw, and larynx (see Fig.~\ref{fig:static_contours}). 
Conceptually, this transformation can be understood as the mapping of high-level vocal tract 
parameters---such as constriction degree and location for consonants, or vocal tract tube shape 
for vowels---onto concrete articulator geometries, akin to the task dynamics approach proposed by 
\citep{SaltzmanMunhall1989}.

The calculation of model articulator positions follows a four-step procedure: (i) computation of 
vocalic articulator positions as a baseline, (ii) computation of primary articulator positions for 
consonantal constrictions, (iii) computation of the lower jaw position as a mediating articulator 
for labial, apical, or dorsal constriction formation, and (iv) adjustment of tongue and lip positions 
in response to the jaw movement.

This stepwise approach reflects the inherent parallelism of vocalic and consonantal articulation 
in fluent speech. During vocalic segments, the articulators are often still moving away from a 
previous consonantal constriction or already transitioning toward the next one. Similarly, 
consonants are never produced in isolation but occur in vocalic context, which means that the 
vocal tract shape during constriction formation is inevitably influenced by both preceding and 
following vowels. In this sense, aspects of dynamic articulation are already embedded within the otherwise static modeling 
framework proposed in this paper.

\textbf{Step 1a:} Vocalic articulator positions for the tongue, jaw, velum, larynx, and lips are 
interpolated based on the high--low and front--back parameters. This interpolation is derived from 
three prototypical vowel shapes corresponding to the cardinal vowels /i/, /a/, and /u/. In the neutral 
case (high--low = 0, front--back = 0), an intermediate position is computed (see 
Fig.~\ref{fig:static_contours}a). For edge cases, high--low = 1 or --1 represents a high or low 
vowel respectively (Fig.~\ref{fig:static_contours}b), and front--back = 1 or --1 indicates a front 
or back vowel respectively (Fig.~\ref{fig:static_contours}c).

\textbf{Step 1b:} The interpolated lip shape is further refined by incorporating a fourth contour: the 
lips and jaw configuration for the rounded front vowel /y/. The spread--rounded parameter controls this 
interpolation, with values of 0 (spread) and 1 (rounded) corresponding to Fig.~\ref{fig:static_contours}d.

\textbf{Step 2:} Based on the current values of the consonantal constriction parameters---i.e., labial 
aperture, tongue tip height, and tongue dorsum height---the model interpolates between the vocalic 
articulator contour (consonantal parameter = 0) and the full constriction or closure (consonantal 
parameter = 1). This applies to lips, tongue tip, and tongue dorsum (Fig.~\ref{fig:static_contours}e--g).

\textbf{Step 3:} The lower jaw position is then interpolated between the value determined by the vocalic 
state (determined by vocalic control parameters) and the consonantal constriction parameter value 
determining  the degree of consonantal constriction. In many cases, this results in a further elevation 
of the jaw beyond its basic vocalic position to support a consonantal target (Fig.~\ref{fig:static_contours}e--g).

\textbf{Step 4:} This jaw-induced elevation (Step 3) in turn leads to secondary adjustments in other 
articulators: (i) In the case of labial constrictions (Fig.~\ref{fig:static_contours}e), the tongue tip 
is elevated; (ii) for dorsal constrictions (Fig.~\ref{fig:static_contours}g), the lower lip is elevated; 
(iii) for apical constrictions (Fig.~\ref{fig:static_contours}f), both the tongue dorsum and lips are 
affected by the jaw’s vertical shift.

\section*{3. Integration of DYNARTmo into a Web App for Speech Therapy and Language Learning}

\begin{figure}[ht]
  \centering
  \includegraphics[width=\textwidth]{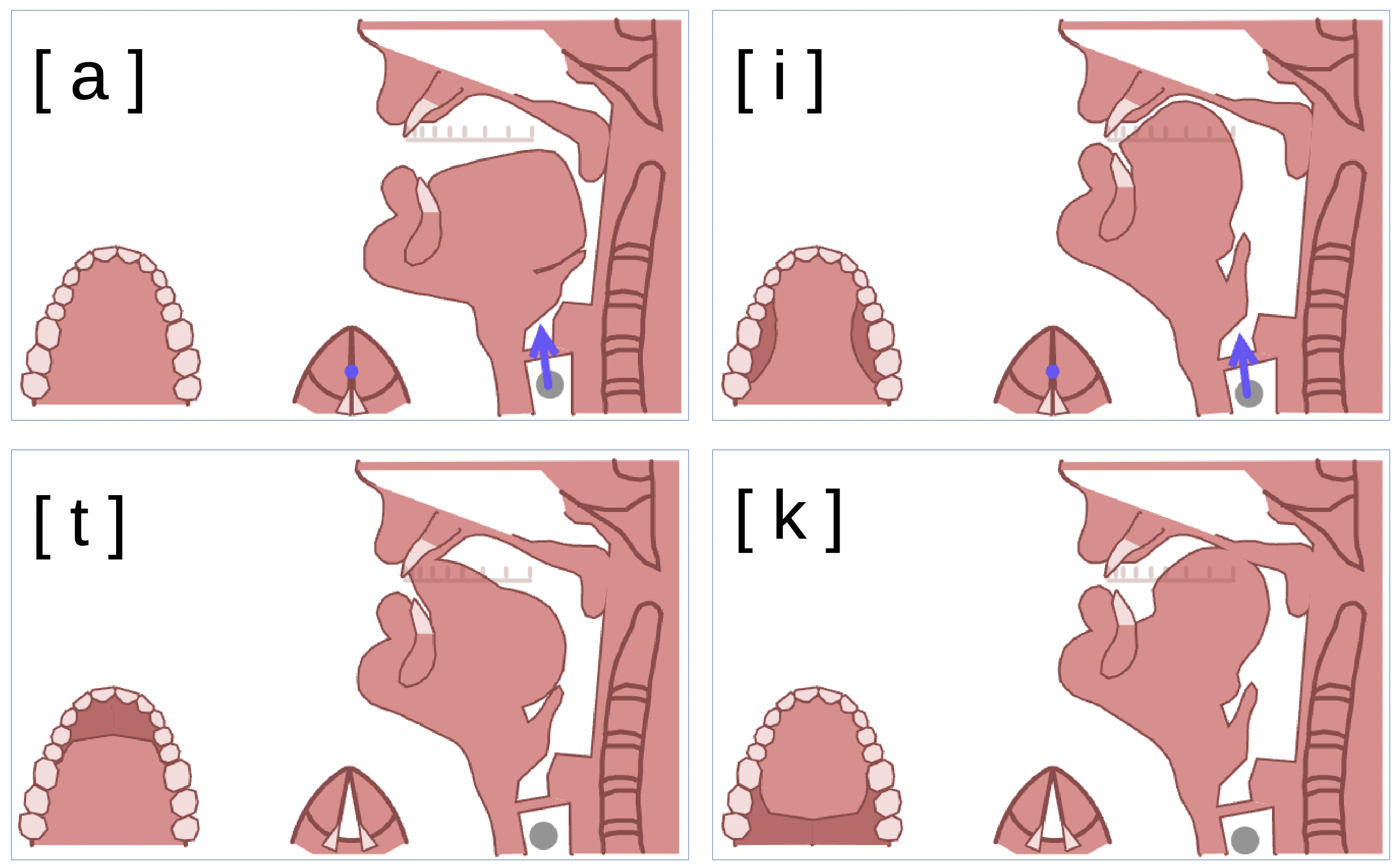}
  \caption{Screenshot from the \textit{SpeechArticulationTrainer} web application, showing the sagittal, 
  glottal, and palatal views for four speech sounds: the vowels [a] and [i], and the consonants [t] 
  and [k]. (The tongue-palate contact pattern for /t/ lacks lateral tongue contact; see Discussion.)}
  \label{fig:multi_view}
\end{figure}

One of the current objectives is to use the model as a visual feedback tool for learning or re-learning 
speech sound production in therapeutic contexts (see Fig.~\ref{fig:multi_view}). Individuals affected 
by speech disorders are often highly motivated to engage with tools that offer visual support to improve 
their articulatory capabilities.

A second goal is to integrate the model into phonetics instruction, for example as part of course 
curricula in linguistics or speech-language pathology (logopaedics).

In addition to displaying the form and displacement of articulators from their rest positions in a 
sagittal plane view (see Fig.~\ref{fig:static_contours}), the \textit{SpeechArticulationTrainer} web 
application also provides  a glottal view (transverse plane, superior perspective), and a 
palatal view (transverse plane, inferior perspective), as illustrated in Figure~\ref{fig:multi_view}.

The glottal view provides visual feedback on the state of the vocal folds: abducted for phonation 
(voiced sounds) versus adducted for the production of voiceless sounds. The palatal view illustrates 
tongue--palate contact regions, which correspond to tactile sensations perceived by the speaker 
during articulation.

Arrows (and dots) indicate the airflow direction, while a grey dot marks the presence of subglottal 
pressure, which is necessary for sound generation.

It should be noted that plosive consonants cannot be fully captured in static images like those shown 
in Figure~\ref{fig:multi_view}. Static frames do not convey the release phase of the constriction or 
the subsequent noise burst characteristic of plosives. For this reason, the application also includes 
the ability to generate animated sequences of articulatory movement patterns---at the syllable, word, 
or short phrase level---enabling more dynamic and comprehensive visualization.

\section*{4. Discussion}

One central goal of this study is to propose a simplified articulatory model that operates without 
the full task-dynamics formalism, while still reflecting key distinctions commonly found in more 
complex frameworks such as the Task Dynamics Approach (\citep{SaltzmanMunhall1989}). Notably, 
our model implicitly includes a distinction between articulator space and tract variable 
space. However, this distinction is not strictly separated in our formulation. For instance, 
articulatory parameters such as tongue dorsum height or velum position are directly mapped to 
constriction properties, thereby bypassing an explicit intermediate representation of tract variables. 
This deliberate simplification serves to make the model computationally tractable and visually 
transparent, especially for educational or diagnostic applications. Nonetheless, the blending of 
articulator-based and functionally-defined parameters might lead to a certain conceptual "blurring" 
that deserves clarification. Future refinements of the model may aim to more 
clearly disentangle these parameter spaces, or at 
least highlight their interplay more explicitly, in order to facilitate theoretical comparisons 
with established models of speech production.

As outlined in Chapter 2, our model differentiates between vocalic and consonantal articulation. 
Moreover, we emphasize there that these two articulatory domains often temporally overlap during 
fluent speech. This overlap is reflected in our approach by describing the interplay between 
vocalic and consonantal gestures within a static articulatory model. This conceptual treatment 
of vowel–consonant interaction can thus be seen as a first step toward 
modeling temporal coordination within articulation. In that sense, these considerations already 
incorporate elements of dynamic articulation into the otherwise static modeling framework presented 
in this paper.

It should be noted that the tongue-palate contact patterns presented in this paper are based on a 
preliminary version of our 3D module for estimating tongue-palate contact. As shown in Fig. 2, the 
current model fails to represent lateral tongue elevation at the molar region in the case of the 
alveolar plosive /t/, resulting in an incorrect apical-only contact. A more complete 3D implementation
—including detailed modeling of the tongue groove—will be incorporated in forthcoming versions of the 
model, covering both plosive and fricative contexts.

\section*{5. Further Work}

Several extensions of the current model are planned for future development and publication. These 
include: (i) the implementation of a virtual 3D model for generating detailed lingual--palatal 
contact patterns, (ii) a comprehensive description of segmental and gestural control mechanisms 
for generating dynamic articulatory movement patterns, and (iii) the integration of aerodynamic 
representation within both the model and the \textit{SpeechArticulationTrainer} web application.

In addition, we aim to reintegrate an articulatory--acoustic module into a neural model framework. 
This will allow DYNARTmo to serve as a component within a large-scale, neurobiologically plausible 
model of speech production and perception \citep{Kroger2023}.

The source code of the current implementation of DYNARTmo is included as Zip-File.

\section*{Acknowledgements}
I am grateful to Georg Heike for introducing me to the fascinating field of articulatory modeling. 
I also wish to thank Catherine Browman, Louis Goldstein, Elliot Saltzman, and Philip Rubin for their 
inspiring discussions during my visit to Haskins Laboratories in New Haven, CT (1994), as well as 
during a phonology conference in Graz, Austria (1992) with Catherine Browman. These interactions 
had a formative influence on the early development of the present model.

\section*{Supplementary Material}
The Python source code used to implement the static visualization routines of DYNARTmo 
(static model part) is provided as a supplementary ZIP archive (\texttt{dynArtMo.zip}) 
attached to this arXiv submission. The archive contains the Jupyter 
notebook \texttt{articulatoryModel.ipynb}, which includes all relevant computational 
routines for generating the sagittal view of vocal tract articulators, as well as a 
subfolder \texttt{data\_sagi} containing the contour data for the reference vocal 
tract shapes. This material allows full reproducibility of the visualizations presented 
in the main text.

\bibliographystyle{apalike}
\bibliography{dynartmo_refs}

\end{document}